\documentclass[conference]{IEEEtran}
\IEEEoverridecommandlockouts
\usepackage{cite}
\usepackage{amsmath,amssymb,amsfonts}
\usepackage{algorithmic}
\usepackage{graphicx}
\usepackage{textcomp}
\usepackage{xcolor}
\usepackage{subcaption}
\def\BibTeX{{\rm B\kern-.05em{\sc i\kern-.025em b}\kern-.08em
		T\kern-.1667em\lower.7ex\hbox{E}\kern-.125emX}}
	
\graphicspath{{figures//}}
\begin{document}
	
	\title{FOGA: Flag Optimization with Genetic Algorithm}
	
	\author{\IEEEauthorblockN{Burak Tağtekin, Berkan Höke, Mert Kutay Sezer and Mahiye Uluyağmur Öztürk}
		\IEEEauthorblockA{Huawei Turkey R\&D Center, Istanbul, Turkey}}
	
	\maketitle
	
	\begin{abstract}
		
		Recently, program autotuning has become very popular especially in embedded systems, when we have limited resources such as computing power and memory where these systems run generally time-critical applications. Compiler optimization space gradually expands with the renewed compiler options and inclusion of new architectures. These advancements bring autotuning even more important position. In this paper, we introduced Flag Optimization with Genetic Algorithm (FOGA) as an autotuning solution for GCC flag optimization. FOGA has two main advantages over the other autotuning approaches: the first one is the hyperparameter tuning of the genetic algorithm (GA), the second one is the maximum iteration parameter to stop when no further improvement occurs. We demonstrated remarkable speedup in the execution time of C++ source codes with the help of optimization flags provided by FOGA when compared to the state of the art framework OpenTuner. 
	\end{abstract}
	
	
	\section{Introduction}
		Vast majority of software applications are developed using a high-level programming language such as C++, Java or C\#. Specific applications that are designed to perform real-time, mission-critical operations or handle low-level memory management on the hardware level are mostly written in C/C++. 
		Such C and C++ codes are passed to the compilers that translate these codes into binary executable files which can be interpreted by computer hardware. Compilers have been employed for 50 years in order to extract platform-specific binary executable from human-readable high-level programming languages\cite{aho1986compilers, hall2009compiler}. Compilers are still developed in terms of optimization so that each code block in a program can be translated into an optimized application. For tuning the compiler flags, some test cases are selected and the resulting performance is analyzed. Afterwards, the performance of the code is analyzed until it obtains the best flag set. Compilers includes three main layers and optimization can be applied to these layers distinctively:
		\begin{itemize}
			\item Frontend
			\item Intermediate Representation
			\item Backend
		\end{itemize}
		Optimizing source codes manually is also an option, but it is highly infeasible, since it is time consuming and probably platform-specific. Instead, compiler optimizations allow us to create optimized binary executable automatically. The scope of this study covers the optimization of intermediate representation of the compiler by enabling/disabling the corresponding optimization flags which yield to significant improvements based on the defined objectives. These objectives could be one of the followings but not limited to; power consumption, execution time (runtime), code size or compile time. The trade-off between these objectives yields to a challenging problem which requires holistic exploration \cite{palermo2005multi}. 
		
		Hence, an efficient approach should be employed in order to solve compiler optimization. Recent advances in artificial intelligence (AI) provides solution to seek optimization parameters using the previously known data. AI is the ideal fit for compiler optimization problem, due to following advantages:
		\begin{itemize}
			\item Once trained, a machine learning model can run fully automatic for previously unseen source code.
			\item Manually optimizing the compiler optimization flags is tedious task, where AI needs less intervention.
			\item Manual optimizations are usually valid for a single architecture or platform but an AI model could be architecture-independent.
		\end{itemize}
		
		 Optimizing the compiler flags to run program faster is also referred as autotuning \cite{datta2008stencil, ansel2011language}. In this study, we employ genetic algorithm approach in order to obtain the best compiler flags for execution time. GCC (GNU Compiler Collection) is the focus of our research for autotuning. Our search space contains 114 compilation flags from version 9.3 which implies that there are $2^{114}$ available combinations for binary flags. Among GCC optimization flags, namely \textbf{-O1}, \textbf{-O2}, \textbf{-O3}, \textbf{-Ofast}. -Ofast and -O3 optimization flags are referred as producing the most optimal flag sets. However, this is not the case for every C/C++ source code and compilation flags should be tweaked for some specific code blocks. To do so, machine learning models are utilized to solve search-space exploration and determining the best flag set. Pre-defined optimization levels of GCC is as follows:
		 
		 \begin{itemize}
		 	\item -O0
		 	\item[] This parameter does not make any optimization on the code, it only allows compiling the code. Main purpose here is to facilitate the debugging process.
		 	\item -O1
		 	\item[] Enables some optimization flags to be compiled by optimizing the code by turning on some optimization flags. Here the compiler attempts to reduce the code size and execution time while keeping the compilation time as low as possible.
		 	\item -O2
		 	\item[] Optimize the code more than -O1. While doing this, the compilation time may be longer since it uses more flags than -O1. In this part, it is necessary to sacrifice memory for higher performance.
		 	\item -O3
		 	\item[] It aims to increase the optimization by turning on flags in -O2, as well as few other optimization flags.
		 \end{itemize}
			
		This study provides a new approach to deal with automatic tuning of the C++ compilation flags by contributing the followings to existing methods:
		\begin{itemize}
			\item Seeking the optimum flag sets is dramatically low compared to other approaches.
			\item In order to find optimum runtime in a sensible time manner we use early stopping strategy.
			\item Hyperparameter fine-tuning of the genetic algorithm with evolutionary algorithms.
		\end{itemize}
	
		The remainder of the paper is organized as follows. First, a thorough literature survey on the problem is provided in Section \ref{sec:rel_work}. Then, methods are introduced to predict compiler optimization flags for minimizing the execution time in Section \ref{sec:method}. Finally, experiments are provided followed by conclusions in Section \ref{sec:exp} and Section \ref{sec:conclusion} respectively.
	
	\section{Related Work}
	\label{sec:rel_work}
		Machine learning models have been used to solve problems from different areas and produces various outputs depending on the problem which includes seeking the best compiler flags or job scheduling and parallel program optimization \cite{wang2018machine}. In this section, we review the machine learning models used for compiler optimization. In general, machine learning model could be either supervised or unsupervised. Supervised learning expects labeled output from user and then reveals the relation between the input and the output such that this relation predicts the output label of the new data which is previously unseen by model. Depending on the prediction output, our supervised model could be either regression if the output has numerical features or classification if the output has categorical features. On the other hand, unsupervised methods do not need labels and mostly used for categorizing the input into several subsets. Recent advances in neural networks and available computing power, deep neural networks and online learning methods gain importance. Deep neural networks consists of many hidden layers where output of each layer is the weighted representation of the input layer depending on the layer type. They need more training data as they become more complex. Additionally, online learning approaches works like trial and error principle and tends to choose best model among all. In this chapter, different approaches of AI are reviewed in terms of compiler flag optimization.
	\subsection{Supervised Learning}
		Regression techniques are widely used for predicting numerical output. In the subject of compiler optimization, Qilin compiler\cite{luk2009qilin} follows an approach so that it predicts the execution time of a program based on the input size for both CPU and GPU. Additionally, another proposed study uses regression techniques to estimate potential speedup of OpenCL programs with specific input \cite{wen2014smart}. Alongside estimating numerical features like execution time and speedup, classification techniques allow us to associate a program with classes implied from the input feature vectors. 
		
		Classification methods are often used for determining parameters. kNN is one of the simple yet effective classification approach used for choosing optimal optimization parameters in previously proposed studies \cite{stephenson2005predicting, micolet2016machine}. kNN algorithm chooses the best optimal flags from programs whose features are closest to the any given program. However, kNN has two major downsides: It gets slower in case of large training data, since it needs to recalculate the distance between the input and training data for each prediction. Secondly, kNN algorithm is not robust to noisy training data and tends to be erroneous, because it simply chooses k nearest neighbors. Decision trees are alternative to kNN for classification tasks. For instance, loop unroll factors are determined by decision trees \cite{monsifrot2002machine, leather2009automatic}, or optimal algorithms are selected for implementation \cite{ding2015autotuning}. Decision trees are easy to interpret and visualize. Nevertheless, they are prone to overfit training data. Random forests have been proposed to overcome overfitting problem \cite{ho1995random}. Lokuciejewski et al.\cite{lokuciejewski2009automatic}, proposed random forests to determine whether to inline a function or not . Also, support vector machines have been used for compiler optimization task in order to classify programs based on their feature vectors \cite{wang2009mapping, zhang2018auto}. 
	\subsection{Unsupervised Learning}
	Unsupervised learning methods such as clustering automatically extract the target label of the input without any intervention instead of labels. Unsupervised learning generally used to reveal distribution of the data or categorize the data into the subgroups. In our specific objective for compiler optimization, works proposed in \cite{perelman2003using, sherwood2002automatically} utilized k-means for clustering program execution path into phase groups. Besides k-means, fast Newman clustering approach \cite{newman2010network} is utilized for categorize functions whose best optimization flags are similar to each other \cite{martins2014exploration}. Unsupervised learning includes dimension reduction techniques as well. Principal component analysis(PCA) and autoencoders are such algorithms to reduce the dimension of feature space \cite{magni2014automatic, vincent2008extracting}. PCA is often employed in order to alleviate the workload of compiler optimization techniques \cite{eeckhout2002workload, chen2010evaluating, ashouri2014bayesian}. On the other hand, some methods have been proposed to extract representative programs from benchmarks \cite{eeckhout2002workload, phansalkar2007analysis}. On the other hand, autoencoders are the artificial neural networks \cite{vincent2008extracting} which compress the features and reconstruct the input from this compressed representation. They are used to make source code feature more compact \cite{gu2016deep}.
	\subsection{Online Learning}
	Evolutionary algorithms (EA) consists of three main categories: genetic algorithms (GA), genetic programming (GP) and stochastic-based search. All of them are utilized to find a good optimization solution in very large search spaces. They work by mimicking the biological evolution to find the best fit for a given problem. Since EA is suitable for large search spaces, it is mostly used for seeking the best compiler flags in previously proposed works \cite{agakov2006using, garciarena2016evolutionary, ballal2015compiler}. The algorithm starts with randomly initialized populations, where each population represents a compiler flag settings. These populations are employed to compile program with various optimization flag sequences and fitness function is generally defined as the execution time of the resulting program. Next population is derived from the current population via mechanisms such as reproduction/crossover and mutation. Compiler flags return lower execution times as these populations evolve by each iteration and algorithm returns the program with the lowest execution time. Agakov et al. \cite{agakov2006using} and Zuluaga et al \cite{zuluaga2012predicting} proposed to combine supervised learning and genetic algorithms by employing supervised model to narrow down the design space. Jantz and Kulkarni \cite{jantz2013exploiting}, proposed a method to prune design space. They pointed out that phase order search space is reduced by 89\% on average.

	\section{Method}
	\label{sec:method}
		
	\subsection{Genetic Algorithm}
	\label{subsec:ga}
	The concept of GA is inspired by the nature,
	weak species are faced with extinction by natural selection. The strong species are more successful in transferring their genes to subsequent generations via reproduction.
	There are parameters that affect the outcome of GA which are selection, crossover and mutation. Crossover, is the most time consuming part of genetic algorithm [17].
	
\begin{figure}[htbp]
		\centering
		\includegraphics[width = 0.8\columnwidth]{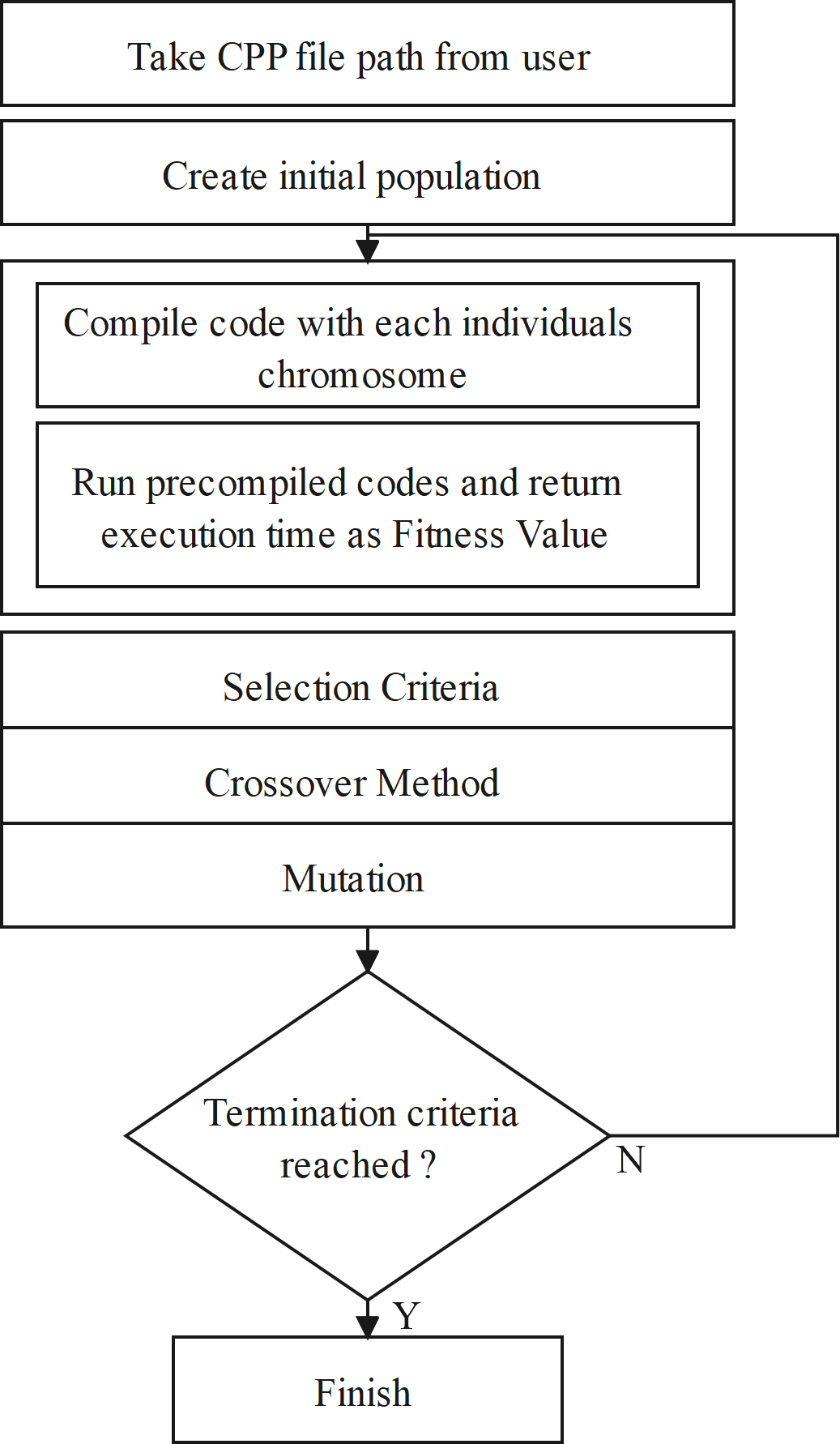}
		\caption{Workflow of genetic algorithm for compiler optimization.}
		\label{fig:workflow}
	\end{figure}

	In this work, we represent compiler optimization flag sets as individuals and each gene of an individual in GA corresponds to a compiler optimization flag. Gene values are binary $\in \{0,1\}$. If the flags are on, gene value is 1, otherwise 0. As a fitness function, we try to optimize the execution time of source codes by choosing the most suitable compiler optimization flags. In other words, the fitness function returns a value proportional to the execution time of the target source code. Hence, if the execution time of the source code is longer, the fitness function will return a larger value or vice versa. Return value of a fitness function represents that individual's chance to survive within a given population (smaller value leads to higher survival ratio). The probability of survival is also related to the selection criteria. Stronger offsprings will be obtained by choosing strong parents for crossover. Considering this in our case, the genes that have little effect on reducing execution time are deleted from the population over time. The value of genes are defined by mostly crossover beside mutation. The mutation rate prevents us from sticking to the local minimas. In our problem, there are $2^{114}$ possible combinations for compiler optimization flags. The genetic algorithm allows us to quickly converge to a solution within such large search sets. Ultimately, when we do not observe any improvement, or when we realize that we do not get better offspring, the genetic algorithm stops and returns the best final individual as a result. Flowchart of the genetic algorithm is shown in Fig. \ref{fig:workflow}.

	\subsection{Tuning Genetic Algorithm}
	\label{subsec:tuning_gen_alg}
	Genetic algorithm has many parameters due to its structure. Population number, mutation rate, crossover type are a few of them. In order to achieve optimum solutions with GA, these parameters should be tuned in a smart way. In previous studies, \cite{sandran2012genetic} only four parameters were tuned on, but we did a very detailed study and wanted to run the genetic algorithm in the most accurate way. In Table \ref{tab:cat_types}, available values for crossover, mutation and selection types are shown. There are $5 \times 2 \times 7$ combinations for categorical variables and 7 different numerical parameters where our aim is tuning these parameters. We had experiments to iteratively compile and run the codes with these parameters. In order to do that, we used OpenTuner to find the best configurations by defining execution time of the best individual (proposed GA parameters of FOGA) as the optimization criteria.
	
		\begin{table}[ht]
		\caption{Available values for crossover, mutation and selection types evaluated to tune genetic algorithm} 
		\label{tab:cat_types}
		\centering 
		\begin{tabular}{|lll|}
			\hline
			\textbf{Crossover Types} & \textbf{Mutation Types} & \textbf{Selection Types} \\
			\hline\hline
			
			\multicolumn{1}{|c|}{One Point} & \multicolumn{1}{|c|}{Gauss by Center} & \multicolumn{1}{|c|}{Fully Random} \\
			\multicolumn{1}{|c|}{Two Point} & \multicolumn{1}{|c|}{Uniform} 		& \multicolumn{1}{|c|}{Roulette} \\
			\multicolumn{1}{|c|}{Uniform} 	& \multicolumn{1}{|c|}{} 				& \multicolumn{1}{|c|}{Stochastic} \\
			\multicolumn{1}{|c|}{Shuffle} 	& \multicolumn{1}{|c|}{} 				& \multicolumn{1}{|c|}{Sigma Scaling} \\
			\multicolumn{1}{|c|}{Segment} 	& \multicolumn{1}{|c|}{} 				& \multicolumn{1}{|c|}{Ranking} \\
			\multicolumn{1}{|c|}{} 			& \multicolumn{1}{|c|}{} 				& \multicolumn{1}{|c|}{Linear Ranking} \\
			\multicolumn{1}{|c|}{} 			& \multicolumn{1}{|c|}{} 				& \multicolumn{1}{|c|}{Tournament} \\ \hline
		\end{tabular}
	\end{table}
	
	
	As a result of the tuning process, we found that the best GA parameters are obtained which seeks compiler optimization flags as quick as possible. Tuned GA parameters are given in Table \ref{tab:tuned_params}. The best result was obtained through segment-based crossover, gauss by center mutation and  linear ranking selection type. In the segment-based crossover, random chosen segments of the parents are used to form an offspring  \cite{altiparmak2009steady}. When compared to other crossover types, segment based crossover has noteworthy diversity on the gene. As a result, the search space contains various flag sets. 
	
	As mutation type gauss by center is used, where a random value which is driven from a normal distribution is added to the chosen gene \cite{back1993overview}. 
	
	In the linear ranking selection concept, the selection probability of an individual is determined by sorting fitness values with equal intervals between adjacent rank \cite{yadav2017comparative}. With linear ranking, individuals with lower fitness values have chance to be chosen. This approach provides to reveal the flags which has not been turned on but may have effect on the execution time.

	\begin{table}[ht]
		\caption{Tuned parameters results} 
		\label{tab:tuned_params}
		\centering 
		\begin{tabular}{|l|l|}
			\hline
			Max Number of Iteration & 100 \\
			Population Size & 277 \\
			Mutation Prob. & 0.287 \\
			Elitism Ratio & 0.147 \\
			Crossover Probability & 0.120 \\
			Parents Portion & 0.689 \\
			Crossover Type & Segment \\
			Mutation Type & Gauss by Center\\
			Selection Type & Linear Ranking \\
			Max Iteration Without Improvement & 45 \\
			\hline
		\end{tabular}
		\label{table:nonlin} 
	\end{table}
		
	\section{Experiments}
	\label{sec:exp}
	Experiments are conducted to observe the runtime differences between autotuned optimization flags by OpenTuner and FOGA as well as the pre-defined optimization options (-O1, -O2, -O3). Three different source codes, namely \textit{tsp\_ga.cpp}, \textit{matrixmutlipy.cpp}, \textit{raytracer.cpp} are used where optimization plays significant role as these source codes mostly rely on linear algebra calculations. We used Ubuntu 20.04 with GCC version 9.3.0 for compiling C++ codes. Our experiments ran on Intel(R) Xeon(R) CPU E5 - 2670 v3 @ 2.30 GHz with 48 total cores.

	\begin{figure*}[htbp]
	\centering
	\begin{subfigure}[b]{0.32\textwidth}
		\centering 
		\includegraphics[width = \textwidth]{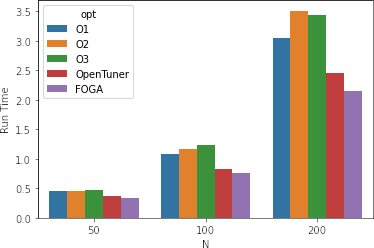}
		\caption{tsp\_ga.cpp}
		\label{subfig:tsp_bar}
	\end{subfigure}
	\begin{subfigure}[b]{0.32\textwidth}
		\centering 
		\includegraphics[width = \textwidth]{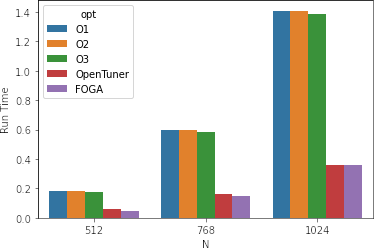}
		\caption{matrixmultiply.cpp}
		\label{subfig:mm_bar}
	\end{subfigure}
	\begin{subfigure}[b]{0.32\textwidth}
		\centering 
		\includegraphics[width = \textwidth]{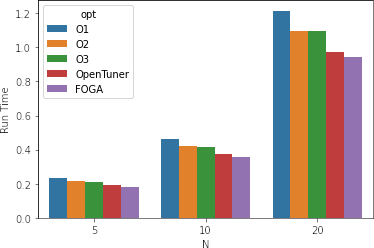}
		\caption{raytracer.cpp}
		\label{subfig:rt_bar}
	\end{subfigure}
	\caption{Comparison of speedup with compiler optimization flags}
	\label{fig:autotune_bar}
\end{figure*}

 	\begin{figure*}[htbp]
	\centering
	\begin{subfigure}[b]{0.32\textwidth}
		\centering
		\begin{tabular}{|lr|}
			\hline
			\textbf{Flags}             & \textbf{Importance} \\ \hline
			funroll-all-loops          & 4.25\%              \\ \hline
			\begin{tabular}[c]{@{}l@{}}funsafe-math \\ -optimizations\end{tabular} & 1.92\%              \\ \hline
			fno-tree-fre               & 1.37\%              \\ \hline
			\begin{tabular}[c]{@{}l@{}} fno-inline \\
				-small-functions\end{tabular} & 1.37\%              \\ \hline
			fno-tree-copy-prop         & 1.23\%              \\ \hline
			111 other flags            & 89.86\%           \\ \hline
		\end{tabular}
		\caption{tsp\_ga.cpp}
		\label{subfig:tsp_imp}
	\end{subfigure}
	\hfill
	\begin{subfigure}[b]{0.32\textwidth}
		\centering
		\begin{tabular}{|lr|}
			\hline
			\textbf{Flags}             & \textbf{Importance} \\ \hline
			\begin{tabular}[c]{@{}l@{}}ftree-loop \\ -vectorize \end{tabular}      & 41.54\%             \\ \hline
			\begin{tabular}[c]{@{}l@{}}funsafe-math \\ -optimizations\end{tabular} & 40.79\%             \\ \hline
			fno-tree-pta               & 0.47\%             \\ \hline
			funroll-all-loops          & 0.39\%              \\ \hline
			fno-dce                    & 0.35\%              \\ \hline
			111 other flags            & 16.46\%            \\ \hline
		\end{tabular}
		\caption{matrixmultiply.cpp}
		\label{subfig:mm_imp}
	\end{subfigure}
	\hfill
	\begin{subfigure}[b]{0.32\textwidth}
		\centering
		\begin{tabular}{|lr|}
			\hline
			\textbf{Flags}              & \textbf{Importance} \\ \hline
			fwrapv                     & 5.31\%              \\ \hline
			fno-unswitch-loops          & 1.64\%              \\ \hline
			fno-forward-propagate       & 1.28\%              \\ \hline
			\begin{tabular}[c]{@{}l@{}} fdelete-null-		  \\ 
				pointer-checks \end{tabular} & 1.28\%         \\ \hline
			fgcse-lm                    & 1.28\%              \\ \hline
			111 other flags             & 89.21\%             \\ \hline
			\multicolumn{1}{c}{} 
		\end{tabular}
		\caption{raytracer.cpp}
		\label{subfig:rt_imp}
	\end{subfigure}
	\caption{Most important flag sets for each source code and their importance percentage values}
	\label{fig:flags_importance}
\end{figure*}

 We compiled the source codes with five different configurations; three of them are the pre-defined optimization options, the fourth configuration is the optimization flag set obtained by OpenTuner\cite{ansel2014opentuner} and the last configuration is the flag set obtained from our proposed method FOGA. 
 
 In Fig. \ref{fig:autotune_bar}, runtime of the pre-defined optimization options, OpenTuner and FOGA are presented, x-axis shows the input size which is denoted as N and y-axis shows the runtime. Also, in Fig \ref{fig:flags_importance} the most important flags for each source code are shown.  We performed one-hot encoding to understand effect of each flag to runtime. Their total speedup factor are rescaled to 100. Fig. \ref{subfig:tsp_bar} represents the execution time of the traveling salesman problem. Execution time of the pre-defined optimization levels are similar, while the execution time of OpenTuner and FOGA decreases as N increases. This is an expected result, since according to Fig. \ref{subfig:tsp_imp} two most important flags, \textit{funroll-all-loops} and \textit{funsafe-math-optimizations} are not included in the default -O1, -O2 or -O3 GCC options. As it is seen in Fig. \ref{subfig:mm_bar}, the margin between the pre-defined optimization levels and the evolutionary methods is wider compared to tsp\_ga. Again, the most important flags are not included in -O1, -O2 or -O3. Moreover, these excluded \textit{tree-loop-vectorize} and \textit{unsafe-math-optimizations} flags are the most important ones for speedup, e.g. responsible for 41.54\% and 40.70\% of the speedup respectively as shown in Fig. \ref{subfig:mm_imp}. Hence, we can deduce that FOGA and OpenTuner successfully find this relation and turn these flags on. On the third example, shown in Fig. \ref{subfig:rt_bar} for \textit{raytracer.cpp}, FOGA and OpenTuner slightly improves the execution time with respect to -O1, -O2, -O3. In this case, improvement ratio is lower, since importance of flags are similar and external flags (those not provided by the pre-defined optimization levels) does not improve the speedup so much for \textit{raytrace.cpp}. In all cases, FOGA outperforms other configurations in terms of execution time regardless of the source code.

	\begin{figure*}[htbp]
		\centering
		\begin{subfigure}[b]{0.32\textwidth}
			\centering 
			\includegraphics[width = \textwidth]{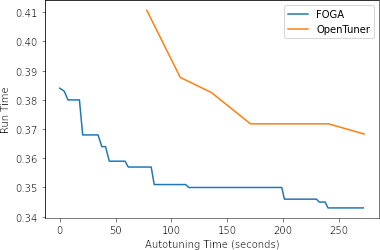}
			\caption{tsp\_ga.cpp}
			\label{subfig:autotune_comp_tsp}
		\end{subfigure}
		\begin{subfigure}[b]{0.32\textwidth}
			\centering 
			\includegraphics[width = \textwidth]{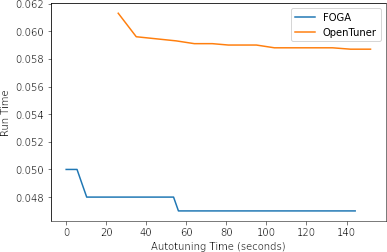}
			\caption{matrixmultiply.cpp}
			\label{subfig:autotune_comp_mm}
		\end{subfigure}
		\begin{subfigure}[b]{0.32\textwidth}
			\centering 
			\includegraphics[width = \textwidth]{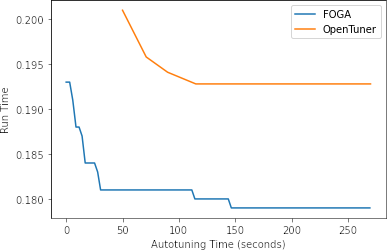}
			\caption{raytracer.cpp}
			\label{subfig:autotune_comp_rt}
		\end{subfigure}
		\caption{Comparison of autotuning speedup for OpenTuner and FOGA.}
		\label{fig:autotune_comp}
	\end{figure*}

	In addition to the execution time of each source code, we also measured the duration of autotuning for reaching the best execution time for each source code. We compared the convergence time of FOGA and OpenTuner to shortest runtime as shown in Fig. \ref{fig:autotune_comp}. For all of the source codes, FOGA has remarkable advantage for both runtime and autotuning time. In Fig. \ref{subfig:autotune_comp_mm} and Fig. \ref{subfig:autotune_comp_rt}, FOGA stops searching whenever it does not provide any improvement after a certain period (see Table \ref{tab:tuned_params}). Note that OpenTuner continues to search for the best flag afterwards, however it could not reach better runtime than FOGA.
	
	\section{Conclusion}
	\label{sec:conclusion}
	Making automatic adjustments at the compiler level is essential for a certain program to work optimally, especially for time-critical applications. With a smart automatic compiler parameter tuning method, it is aimed to increase performance by adjusting the compiler optimization flags. Sometimes turning off a compiler flag that is enabled by default increases the performance or in some cases activating a specific parameter enhances the efficiency. The mission of a autotuning method is to find the best possible parameters. In this study, we proposed FOGA which automatically tunes the compiler optimization flags with the help of genetic algorithm. As a first step, we tuned the parameters of genetic algorithm via OpenTuner by iteratively search among many combinations. Secondly, we showed that FOGA surpassed OpenTuner in terms of reaching the minimum execution time of a program. Additionally, we limited the iteration count if there is no improvement on the execution time for greater than a \textit{Maximum Iteration without Improvement} parameter. These improvements allow us to reach shorter execution times in autotuning time with respect to the pre-defined optimization levels and OpenTuner. 
	
	As a future work, we plan to implement genetic algorithm that posses customized mutation and crossover operations. These customized operations are employed to keep alive important genes for the next generations. This approach narrows down the search space as individuals evolved and enable to find the best flags sooner.
		
	\bibliographystyle{IEEEtran}
	\bibliography{refs}

\begin{thebibliography}{10}
\providecommand{\url}[1]{#1}
\csname url@samestyle\endcsname
\providecommand{\newblock}{\relax}
\providecommand{\bibinfo}[2]{#2}
\providecommand{\BIBentrySTDinterwordspacing}{\spaceskip=0pt\relax}
\providecommand{\BIBentryALTinterwordstretchfactor}{4}
\providecommand{\BIBentryALTinterwordspacing}{\spaceskip=\fontdimen2\font plus
\BIBentryALTinterwordstretchfactor\fontdimen3\font minus
  \fontdimen4\font\relax}
\providecommand{\BIBforeignlanguage}[2]{{%
\expandafter\ifx\csname l@#1\endcsname\relax
\typeout{** WARNING: IEEEtran.bst: No hyphenation pattern has been}%
\typeout{** loaded for the language `#1'. Using the pattern for}%
\typeout{** the default language instead.}%
\else
\language=\csname l@#1\endcsname
\fi
#2}}
\providecommand{\BIBdecl}{\relax}
\BIBdecl

\bibitem{aho1986compilers}
A.~V. Aho, R.~Sethi, and J.~D. Ullman, ``Compilers, principles, techniques,''
  \emph{Addison wesley}, vol.~7, no.~8, p.~9, 1986.

\bibitem{hall2009compiler}
M.~Hall, D.~Padua, and K.~Pingali, ``Compiler research: the next 50 years,''
  \emph{Communications of the ACM}, vol.~52, no.~2, pp. 60--67, 2009.

\bibitem{palermo2005multi}
G.~Palermo, C.~Silvano, and V.~Zaccaria, ``Multi-objective design space
  exploration of embedded systems,'' \emph{Journal of Embedded Computing},
  vol.~1, no.~3, pp. 305--316, 2005.

\bibitem{datta2008stencil}
K.~Datta, M.~Murphy, V.~Volkov, S.~Williams, J.~Carter, L.~Oliker,
  D.~Patterson, J.~Shalf, and K.~Yelick, ``Stencil computation optimization and
  auto-tuning on state-of-the-art multicore architectures,'' in \emph{SC'08:
  Proceedings of the 2008 ACM/IEEE conference on Supercomputing}.\hskip 1em
  plus 0.5em minus 0.4em\relax IEEE, 2008, pp. 1--12.

\bibitem{ansel2011language}
J.~Ansel, Y.~L. Wong, C.~Chan, M.~Olszewski, A.~Edelman, and S.~Amarasinghe,
  ``Language and compiler support for auto-tuning variable-accuracy
  algorithms,'' in \emph{International Symposium on Code Generation and
  Optimization (CGO 2011)}.\hskip 1em plus 0.5em minus 0.4em\relax IEEE, 2011,
  pp. 85--96.

\bibitem{wang2018machine}
Z.~Wang and M.~O’Boyle, ``Machine learning in compiler optimization,''
  \emph{Proceedings of the IEEE}, vol. 106, no.~11, pp. 1879--1901, 2018.

\bibitem{luk2009qilin}
C.-K. Luk, S.~Hong, and H.~Kim, ``Qilin: exploiting parallelism on
  heterogeneous multiprocessors with adaptive mapping,'' in \emph{2009 42nd
  Annual IEEE/ACM International Symposium on Microarchitecture (MICRO)}.\hskip
  1em plus 0.5em minus 0.4em\relax IEEE, 2009, pp. 45--55.

\bibitem{wen2014smart}
Y.~Wen, Z.~Wang, and M.~F. O'boyle, ``Smart multi-task scheduling for opencl
  programs on cpu/gpu heterogeneous platforms,'' in \emph{2014 21st
  International conference on high performance computing (HiPC)}.\hskip 1em
  plus 0.5em minus 0.4em\relax IEEE, 2014, pp. 1--10.

\bibitem{stephenson2005predicting}
M.~Stephenson and S.~Amarasinghe, ``Predicting unroll factors using supervised
  classification,'' in \emph{International symposium on code generation and
  optimization}.\hskip 1em plus 0.5em minus 0.4em\relax IEEE, 2005, pp.
  123--134.

\bibitem{micolet2016machine}
P.-J. Micolet, A.~Smith, and C.~Dubach, ``A machine learning approach to
  mapping streaming workloads to dynamic multicore processors,'' in
  \emph{Proceedings of the 17th ACM SIGPLAN/SIGBED Conference on Languages,
  Compilers, Tools, and Theory for Embedded Systems}, 2016, pp. 113--122.

\bibitem{monsifrot2002machine}
A.~Monsifrot, F.~Bodin, and R.~Quiniou, ``A machine learning approach to
  automatic production of compiler heuristics,'' in \emph{International
  conference on artificial intelligence: methodology, systems, and
  applications}.\hskip 1em plus 0.5em minus 0.4em\relax Springer, 2002, pp.
  41--50.

\bibitem{leather2009automatic}
H.~Leather, E.~Bonilla, and M.~O'Boyle, ``Automatic feature generation for
  machine learning based optimizing compilation,'' in \emph{2009 International
  Symposium on Code Generation and Optimization}.\hskip 1em plus 0.5em minus
  0.4em\relax IEEE, 2009, pp. 81--91.

\bibitem{ding2015autotuning}
Y.~Ding, J.~Ansel, K.~Veeramachaneni, X.~Shen, U.-M. O’Reilly, and
  S.~Amarasinghe, ``Autotuning algorithmic choice for input sensitivity,''
  \emph{ACM SIGPLAN Notices}, vol.~50, no.~6, pp. 379--390, 2015.

\bibitem{ho1995random}
T.~K. Ho, ``Random decision forests,'' in \emph{Proceedings of 3rd
  international conference on document analysis and recognition}, vol.~1.\hskip
  1em plus 0.5em minus 0.4em\relax IEEE, 1995, pp. 278--282.

\bibitem{lokuciejewski2009automatic}
P.~Lokuciejewski, F.~Gedikli, P.~Marwedel, and K.~Morik, ``Automatic wcet
  reduction by machine learning based heuristics for function inlining,'' in
  \emph{3rd workshop on statistical and machine learning approaches to
  architectures and compilation (SMART)}, 2009, pp. 1--15.

\bibitem{wang2009mapping}
Z.~Wang and M.~F. O'Boyle, ``Mapping parallelism to multi-cores: a machine
  learning based approach,'' in \emph{Proceedings of the 14th ACM SIGPLAN
  symposium on Principles and practice of parallel programming}, 2009, pp.
  75--84.

\bibitem{zhang2018auto}
P.~Zhang, J.~Fang, T.~Tang, C.~Yang, and Z.~Wang, ``Auto-tuning streamed
  applications on intel xeon phi,'' in \emph{2018 IEEE International Parallel
  and Distributed Processing Symposium (IPDPS)}.\hskip 1em plus 0.5em minus
  0.4em\relax IEEE, 2018, pp. 515--525.

\bibitem{perelman2003using}
E.~Perelman, G.~Hamerly, M.~Van~Biesbrouck, T.~Sherwood, and B.~Calder, ``Using
  simpoint for accurate and efficient simulation,'' \emph{ACM SIGMETRICS
  Performance Evaluation Review}, vol.~31, no.~1, pp. 318--319, 2003.

\bibitem{sherwood2002automatically}
T.~Sherwood, E.~Perelman, G.~Hamerly, and B.~Calder, ``Automatically
  characterizing large scale program behavior,'' \emph{ACM SIGPLAN Notices},
  vol.~37, no.~10, pp. 45--57, 2002.

\bibitem{newman2010network}
M.~Newman, \emph{Networks: An Introduction}.\hskip 1em plus 0.5em minus
  0.4em\relax New York, NY, USA: Oxford University Press, 2010.

\bibitem{martins2014exploration}
L.~G. Martins, R.~Nobre, A.~C. Delbem, E.~Marques, and J.~M. Cardoso,
  ``Exploration of compiler optimization sequences using clustering-based
  selection,'' in \emph{Proceedings of the 2014 SIGPLAN/SIGBED conference on
  Languages, compilers and tools for embedded systems}, 2014, pp. 63--72.

\bibitem{magni2014automatic}
A.~Magni, C.~Dubach, and M.~O'Boyle, ``Automatic optimization of
  thread-coarsening for graphics processors,'' in \emph{Proceedings of the 23rd
  international conference on Parallel architectures and compilation}, 2014,
  pp. 455--466.

\bibitem{vincent2008extracting}
P.~Vincent, H.~Larochelle, Y.~Bengio, and P.-A. Manzagol, ``Extracting and
  composing robust features with denoising autoencoders,'' in \emph{Proceedings
  of the 25th international conference on Machine learning}, 2008, pp.
  1096--1103.

\bibitem{eeckhout2002workload}
L.~Eeckhout, H.~Vandierendonck, and K.~De~Bosschere, ``Workload design:
  Selecting representative program-input pairs,'' in \emph{Proceedings.
  International Conference on Parallel Architectures and Compilation
  Techniques}.\hskip 1em plus 0.5em minus 0.4em\relax IEEE, 2002, pp. 83--94.

\bibitem{chen2010evaluating}
Y.~Chen, Y.~Huang, L.~Eeckhout, G.~Fursin, L.~Peng, O.~Temam, and C.~Wu,
  ``Evaluating iterative optimization across 1000 datasets,'' in
  \emph{Proceedings of the 31st ACM SIGPLAN Conference on Programming Language
  Design and Implementation}, 2010, pp. 448--459.

\bibitem{ashouri2014bayesian}
A.~H. Ashouri, G.~Mariani, G.~Palermo, and C.~Silvano, ``A bayesian network
  approach for compiler auto-tuning for embedded processors,'' in \emph{2014
  IEEE 12th Symposium on Embedded Systems for Real-time Multimedia
  (ESTIMedia)}.\hskip 1em plus 0.5em minus 0.4em\relax IEEE, 2014, pp. 90--97.

\bibitem{phansalkar2007analysis}
A.~Phansalkar, A.~Joshi, and L.~K. John, ``Analysis of redundancy and
  application balance in the spec cpu2006 benchmark suite,'' in
  \emph{Proceedings of the 34th annual International Symposium on Computer
  architecture}, 2007, pp. 412--423.

\bibitem{gu2016deep}
X.~Gu, H.~Zhang, D.~Zhang, and S.~Kim, ``Deep api learning,'' in
  \emph{Proceedings of the 2016 24th ACM SIGSOFT International Symposium on
  Foundations of Software Engineering}, 2016, pp. 631--642.

\bibitem{agakov2006using}
F.~Agakov, E.~Bonilla, J.~Cavazos, B.~Franke, G.~Fursin, M.~F. O'Boyle,
  J.~Thomson, M.~Toussaint, and C.~K. Williams, ``Using machine learning to
  focus iterative optimization,'' in \emph{International Symposium on Code
  Generation and Optimization (CGO'06)}.\hskip 1em plus 0.5em minus 0.4em\relax
  IEEE, 2006, pp. 11--pp.

\bibitem{garciarena2016evolutionary}
U.~Garciarena and R.~Santana, ``Evolutionary optimization of compiler flag
  selection by learning and exploiting flags interactions,'' in
  \emph{Proceedings of the 2016 on Genetic and Evolutionary Computation
  Conference Companion}, 2016, pp. 1159--1166.

\bibitem{ballal2015compiler}
P.~A. Ballal, H.~Sarojadevi, and P.~Harsha, ``Compiler optimization: A genetic
  algorithm approach,'' \emph{International Journal of Computer Applications},
  vol. 112, no.~10, 2015.

\bibitem{zuluaga2012predicting}
M.~Zuluaga, E.~Bonilla, and N.~Topham, ``Predicting best design trade-offs: A
  case study in processor customization,'' in \emph{2012 Design, Automation \&
  Test in Europe Conference \& Exhibition (DATE)}.\hskip 1em plus 0.5em minus
  0.4em\relax IEEE, 2012, pp. 1030--1035.

\bibitem{jantz2013exploiting}
M.~R. Jantz and P.~A. Kulkarni, ``Exploiting phase inter-dependencies for
  faster iterative compiler optimization phase order searches,'' in \emph{2013
  International Conference on Compilers, Architecture and Synthesis for
  Embedded Systems (CASES)}.\hskip 1em plus 0.5em minus 0.4em\relax IEEE, 2013,
  pp. 1--10.

\bibitem{sandran2012genetic}
T.~Sandran, M.~N.~B. Zakaria, and A.~J. Pal, ``A genetic algorithm approach
  towards compiler flag selection based on compilation and execution
  duration,'' in \emph{2012 International Conference on Computer \& Information
  Science (ICCIS)}, vol.~1.\hskip 1em plus 0.5em minus 0.4em\relax IEEE, 2012,
  pp. 270--274.

\bibitem{altiparmak2009steady}
F.~Altiparmak, M.~Gen, L.~Lin, and I.~Karaoglan, ``A steady-state genetic
  algorithm for multi-product supply chain network design,'' \emph{Computers \&
  Industrial Engineering}, vol.~56, no.~2, pp. 521--537, 2009.

\bibitem{back1993overview}
T.~B{\"a}ck and H.-P. Schwefel, ``An overview of evolutionary algorithms for
  parameter optimization,'' \emph{Evolutionary computation}, vol.~1, no.~1, pp.
  1--23, 1993.

\bibitem{yadav2017comparative}
S.~L. Yadav and A.~Sohal, ``Comparative study of different selection techniques
  in genetic algorithm,'' \emph{Journal Homepage: http://www. ijesm. co. in},
  vol.~6, no.~3, 2017.

\bibitem{ansel2014opentuner}
J.~Ansel, S.~Kamil, K.~Veeramachaneni, J.~Ragan-Kelley, J.~Bosboom, U.-M.
  O'Reilly, and S.~Amarasinghe, ``Opentuner: An extensible framework for
  program autotuning,'' in \emph{Proceedings of the 23rd international
  conference on Parallel architectures and compilation}, 2014, pp. 303--316.

\end{thebibliography}
	
\end{document}